\documentclass{article} 
\usepackage{iclr2027_conference,times}

\usepackage{amsmath,amsfonts,bm}

\def\eqref#1{equation~\ref{#1}}

\def\1{\bm{1}}

\DeclareMathAlphabet{\mathsfit}{\encodingdefault}{\sfdefault}{m}{sl}
\SetMathAlphabet{\mathsfit}{bold}{\encodingdefault}{\sfdefault}{bx}{n}

\usepackage{hyperref}
\usepackage{url}
\usepackage{booktabs}
\usepackage{graphicx} 
\usepackage{multirow}
\usepackage{enumitem}
\usepackage[most]{tcolorbox}

\title{GameDirector: Decoupling Gameplay Logic from Rendering for Player-Configurable Game World Models}

\author{
Zijun Lin$^{1,2,4,\ast,\dagger}$,
Zhiyang Deng$^{1,3,\ast,\dagger}$,
Yuzhe Wu$^{1,3,\dagger}$,
Bihan Wen$^{2}$,
Yeying Jin$^{1,3,\S}$
\\[0.5em]
$^{1}$Tencent \\
$^{2}$Nanyang Technological University \\
$^{3}$National University of Singapore \\
$^{4}$Centre for Frontier AI Research, A*STAR
}

\iclrfinalcopy 

\begin{document}

\maketitle

\fancyhead[L]{Preprint.}

\begingroup

\renewcommand{\thefootnote}{\ensuremath{\ast}}
\footnotetext{Equal contribution.}

\renewcommand{\thefootnote}{\S}
\footnotetext{Corresponding author and project leader.}

\renewcommand{\thefootnote}{\ensuremath{\dagger}}
\footnotetext{This work was completed during research internships at Tencent under the supervision of Yeying Jin.}

\endgroup

\begin{abstract}
Recent game world models support realistic visual simulation and interactive gameplay based on player inputs. However, they typically learn environment dynamics from pixel-level supervision, jointly modeling perception, memory, state transitions, and rendering within an end-to-end framework. While this design enables open-ended, action-controllable generation, it still falls short of delivering a complete gameplay experience. Games are governed by explicit mechanics, such as health deduction, skill activation, combat rules, and termination conditions. These mechanics depend on precise and consistent state transitions that generative models alone cannot reliably enforce. In contrast, game engines can guarantee such mechanics through hard-coded rules, but provide limited flexibility for player-driven creation. To bridge these paradigms, we introduce \textbf{GameDirector}, the first agentic framework that decouples rule-based gameplay logic from visual rendering. Given player-defined configurations, the framework acts as an intelligent director that interprets visual observations, updates game states, tactically controls NPCs, and enforces gameplay rules. It then translates these decisions into text prompts that guide the video world model to render the resulting gameplay. This separation allows players to configure characters, states, and rules much like a game developer while preserving coherent game mechanics. Experiments on three games, using data collected by our automated gameplay agent, show that GameDirector achieves accurate state tracking, reliable rule following, and improves boss action quality by more than 39.9\% over various end-to-end game world model settings. Overall, by externalizing player-controllable game logic, GameDirector establishes an effective middle ground between hard-coded simulation and generative modeling, enabling more flexible and closed-loop gameplay experiences. Project Page: \href{https://jimntu.github.io/gamedirector/}{\textcolor{blue}{https://jimntu.github.io/gamedirector/}}

\end{abstract}

\vspace{-11pt}

\section{Introduction}
Recent advances in world models have demonstrated promising capabilities in generating visually realistic and interactive environments that respond dynamically to user actions \citep{bruce2024genie,shen2026lyra,team2026dreamx}. This naturally motivates their application to games, where players continuously interact with the environment through keyboard and mouse inputs, making games an ideal testbed for action-conditioned world modeling. Recent game world models have further extended this capability to support combat with non-player characters (NPCs) \citep{zhu2026incantation}, dynamic environment changes \citep{tong2026scope}, and multiplayer scenarios \citep{wu2026multiworld}. Compared with traditional games built on hard-coded engines, this generative paradigm offers greater flexibility. In particular, it allows players to customize characters, scenarios, and gameplay settings, opening new possibilities for player-driven game experiences.

Despite this flexibility, generative modeling remains insufficient to fully replace conventional game engines. Games are not governed by open-ended interaction alone; they are fundamentally structured by explicit mechanics and constraints that regulate gameplay \citep{li2026pixels, lin2026stateplay}. For instance, a game may terminate when a character’s health reaches zero, a skill may only be activated once specific conditions are satisfied, and bosses are expected to act according to the evolving game state. Such mechanics are essential for maintaining coherent state transitions, enforcing gameplay rules, and providing meaningful challenges to the player. 

\begin{figure*}[t]
\centering
    \includegraphics[width=0.99\textwidth]{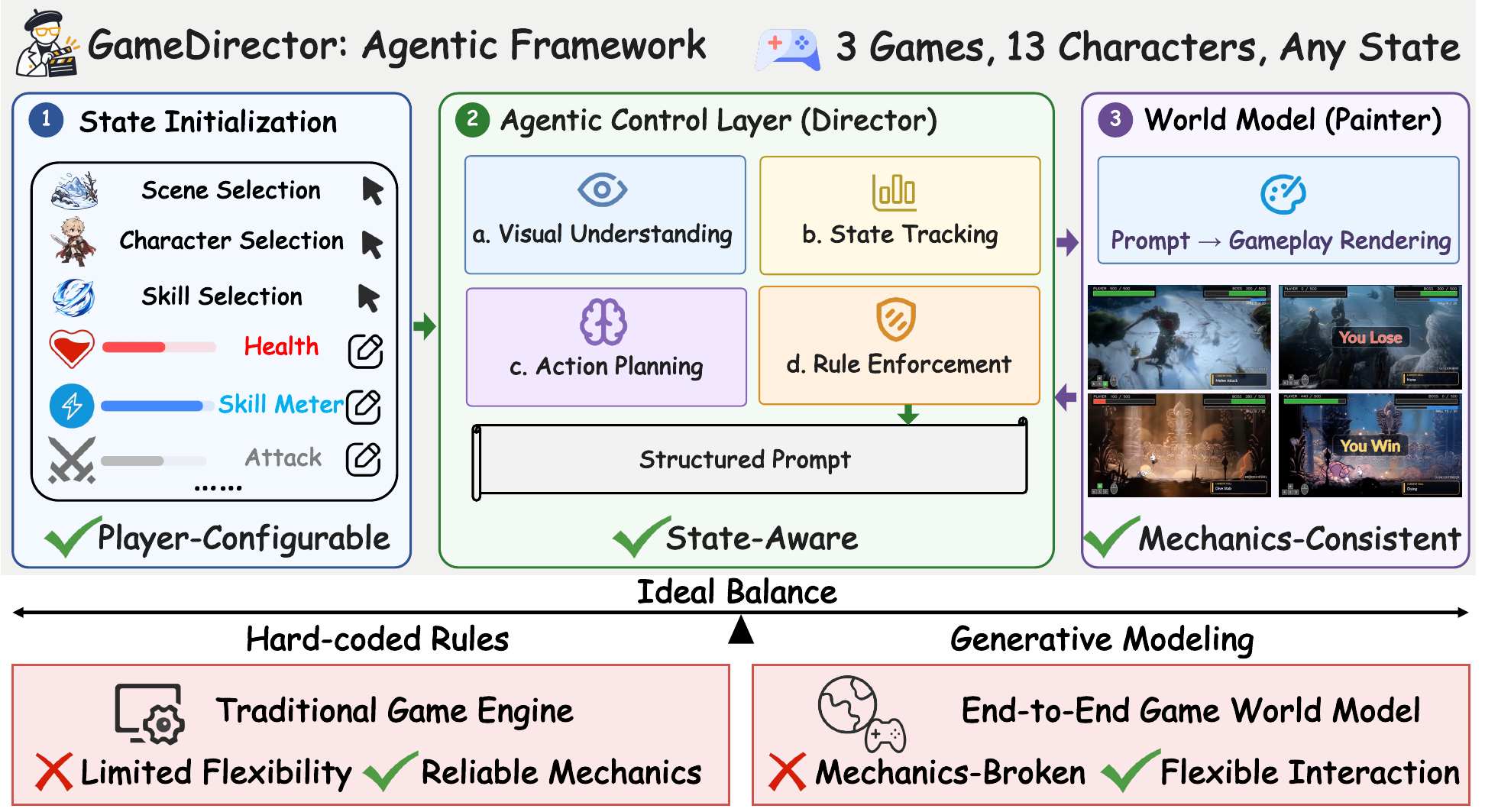}
    \caption{
    Overview of \textbf{GameDirector}. Our framework bridges hard-coded rules and generative modeling to enable player-configurable, state-aware, and mechanics-consistent gameplay.
    }
    \label{fig:teaser}
\end{figure*}

However, existing game world models either largely neglect explicit state and rule modeling or entangle these capabilities with visual generation within a unified generative framework \citep{li2026wildworld,lin2026stateplay}. Although such designs may be effective under relatively simple and fixed game configurations, they are difficult to extend to diverse player-defined settings in which characters, states, skills, and gameplay rules can vary substantially. This motivates a key question: \textit{How can we construct game environments that retain the flexibility of player-configurable generation while consistently satisfying explicit gameplay mechanics and constraints?}

To address this gap, we propose \textbf{GameDirector}, the first agentic framework that decouples rule-based game logic from visual rendering. As illustrated in Fig. \ref{fig:teaser}, instead of relying on a single generative model to jointly reason about gameplay dynamics and synthesize visual content, GameDirector externalizes game control into four explicit components.\textbf{Visual Understanding} interprets the generated observations to extract key gameplay signals, including combat situations and hit events. \textbf{State Tracking} maintains and updates explicit states such as health points and skill meters based on the detected signals. \textbf{Action Planning} uses a 2B language model to select the next boss action from the spatial context, while \textbf{Rule Enforcement} ensures that actions and state transitions follow predefined mechanics, including skill activation and termination rules. These decisions are converted into structured prompts for the video world model, which focuses on faithfully rendering the resulting gameplay. Importantly, this decoupled design enables players to personalize their gameplay experience by specifying preferred characters and combat configurations, thereby customizing both the appearance and difficulty of their opponents like game developers. Such player-defined control is difficult to achieve with a monolithic world model, but becomes naturally supported through our agentic framework.

To evaluate our framework, we develop an automated gameplay agent that collects video recordings with synchronized internal game states across three games. Compared with representative settings used in existing end-to-end game world models, our agentic framework achieves highly reliable state tracking and 98.5\% mechanics fidelity over 1-minute rollouts. It also maintains over 90\% accuracy for attack detection and spatial understanding across the three games, improving boss decision quality by more than 39.9\% and reaching up to 93.0\% accuracy.

Overall, our contributions are summarized as follows:
\vspace{-10pt}
\begin{itemize}
    \item We propose \textbf{GameDirector}, the first agentic framework that combines the rule consistency of game engines with the flexibility of generative models, preserving reliable game mechanics while supporting open-ended visual generation.

    \item We decouple visual understanding, state tracking, action planning, and rule enforcement from rendering through an agentic control layer, enabling player-configurable gameplay with consistent mechanics and adaptive boss behavior over long-horizon generation.

    \item We construct a state-annotated gameplay dataset spanning three games and show that \textbf{GameDirector} substantially outperforms end-to-end game world model settings, achieving over 98.5\% mechanics fidelity and improving boss decision quality by more than 39.9\%.
\end{itemize}

\section{Related Work}
\subsection{Interactive World Models}
Recent advances in video generation have enabled interactive world models that produce high-quality visual rollouts under real-time user control \citep{luo2026aigamesfoundationmodel, yin2026alaya, mao2026yume1, zhu2026sana, team2026inspatio, chen2026h3}.
ReactiveGWM \citep{wang2026reactivegwm} models player--NPC interactions, while SCOPE \citep{tong2026scope} improves fine-grained responsiveness and cross-game generalization in FPS environments.
Incantation \citep{zhu2026incantation} introduces natural language as a unified interface for fine-grained multi-entity control, while Matrix-Game \citep{he2025matrix} and LingBot-World \citep{team2026advancing} demonstrate long-horizon, real-time interactive generation.
Despite increasingly realistic and responsive generation, these methods mainly focus on visual dynamics and action controllability.
Complete gameplay additionally requires explicit internal states and mechanics, including health evolution, skill activation, and termination conditions, which visual generation alone cannot guarantee.

\subsection{State-Aware Game World Models}
Recent works have begun to explicitly model gameplay states beyond action-conditioned generation.
WildWorld \citep{li2026wildworld} provides large-scale gameplay data with synchronized states, actions, and observations, highlighting state consistency for long-horizon generation.
StatePlay \citep{lin2026stateplay} jointly predicts internal states and visual content, using variables such as health points and skill meters to improve mechanics consistency.
Marionette \citep{meng2026marionette} predicts an articulated 3D world state before rendering geometry and appearance, while WorldMind \citep{deng2026worldmind} decouples visual understanding and state-grounded decision making for responsive NPC behavior.
However, these methods still rely on learned state prediction or mainly use states for generation and decision making.
GameDirector instead maintains explicit states and enforces rules through an agentic control layer, supporting reliable mechanics and player-defined configurations in closed-loop gameplay.

\subsection{Agentic World Models}
A concurrent line of work separates world evolution from neural rendering through external reasoning or executable structures.
Code World Model \citep{chen2026code} uses a coding agent to maintain persistent states and executable world dynamics, which are converted into visual constraints for video generation.
Magpie \citep{zhan2026magpie} similarly separates gameplay execution from generative rendering, but relies on a conventional game engine for states and rules.
Programmable World Model \citep{huang2026programmable} translates natural-language specifications into executable state-transition programs and represents world states with state-augmented 3D bounding boxes.
LingBot-World 2.0 \citep{gao2026infinite} further introduces pilot and director agents for character behavior and scene evolution.
These works demonstrate structured control around generative world models, but do not jointly integrate visual understanding, explicit state tracking, adaptive NPC planning, and rule enforcement for closed-loop, player-configurable gameplay.
GameDirector unifies these capabilities in a single agentic control layer while leaving the world model focused on visual rendering.

\vspace{-6pt}
\section{GameDirector}
\vspace{-5pt}
\label{sec:gamedirector}

Generating playable game environments requires both flexible visual synthesis and reliable state-dependent mechanics. To achieve this, we propose \textbf{GameDirector}, an agentic framework that connects configurable game mechanics with a generative world model. Using the dataset collected by our automated gameplay agent (Sec.~\ref{subsec:dataset}), we first introduce player-configurable initialization, where players specify the characters, scene, and initial combat states (Sec.~\ref{subsec:gamedirector_initialization}). We then present an agentic control layer that acts as the \emph{director}, integrating visual understanding, state tracking, action planning, and rule enforcement to select valid actions based on a compact state representation and the current game context (Sec.~\ref{subsec:gamedirector_control}). Finally, we introduce controllable generation, in which the world model acts as the \emph{painter}, faithfully translating structured action prompts produced by the control layer into gameplay frames (Sec.~\ref{subsec:gamedirector_generation}). The generated frames are fed back to the control layer for subsequent state updates and action decisions, closing the loop between explicit game mechanics and visual generation to enable a complete gameplay experience.

\begin{figure*}[t]
\centering
    \includegraphics[width=\textwidth]{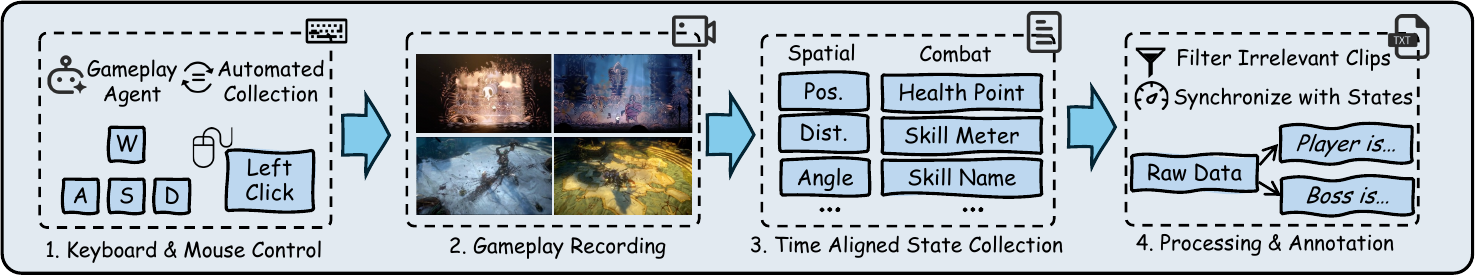}
    \caption{Data Collection Pipeline. An automated gameplay agent collects keyboard and mouse inputs, gameplay recordings, and temporally aligned state information across three games, followed by filtering, synchronization, and annotation to construct the final dataset.
    }
    \label{fig:data}
\end{figure*}

\subsection{Dataset Construction}
\label{subsec:dataset}
GameDirector requires supervision beyond standard video--action pairs to support both visual generation and explicit gameplay reasoning. As shown in Fig.~\ref{fig:data}, we develop an automated gameplay agent that collects game videos together with synchronized combat and spatial states, including health points, skill meters, executed skills, character positions, facing directions, relative angles, and distances. Irrelevant segments are removed, and raw control and action logs are converted into structured prompts on a shared timeline, providing supervision for AttackNet, SituationNet, and action-conditioned generation. Using this pipeline, we construct a dataset across three games: \emph{No Rest for the Wicked}, \emph{Vampire}, and \emph{Hollow Knight}, denoted as \textbf{Game N}, \textbf{Game V}, and \textbf{Game H}. It contains 51,786 clips covering 13 characters (5 playable characters and 8 bosses), with Game V, Game N, and Game H accounting for 36.0\%, 43.8\%, and 20.1\%, respectively. We reserve 300 clips for evaluation, with 100 per game. Additional dataset details are provided in Appendix ~\ref{app:dataset}.

\subsection{Player-Configurable Initialization}
\label{subsec:gamedirector_initialization}

GameDirector begins with player-configurable initialization, allowing players to select their preferred scene and characters, configure combat parameters, and assign to the boss skills originally associated with other bosses. These choices make each match distinct and personalized.

The selected scene and characters determine an initial frame $\mathbf{F}_0$. We denote the initial health points of the player and boss by $h_0^p$ and $h_0^b$, their attack strengths by $\alpha^p$ and $\alpha^b$, and the initial boss skill-meter value by $m_0^b$. The selected skill set is defined as $\mathcal{C}=\{c_i\}_{i=1}^{N}$, where $c_i$ denotes the $i$-th skill and specifies its description, effective range, execution duration, skill-meter threshold, and cooldown time. All these parameters can be freely configured by the player. The complete initialization is represented as
\begin{equation}
    \mathcal{I}_0
    =
    \big[
        \mathbf{F}_0,h_0^p,h_0^b,\alpha^p,\alpha^b,m_0^b,\mathcal{C}
    \big].
\end{equation}

Here, $\mathbf{F}_0$ is provided to the world model as its initial visual condition, while the combat parameters and skill configurations are passed to the agentic control layer, which maintains and updates the corresponding game states throughout gameplay.

Importantly, player configuration is not limited to the variables listed in $\mathcal{I}_0$. Players may also modify the attributes of individual skills in $\mathcal{C}$, such as their cooldown times and skill-meter thresholds, or introduce additional game-specific variables. Because the agentic control layer introduced in Sec.~\ref{subsec:gamedirector_control} operates directly on these explicit configurations, it can accommodate such changes without modifying the world model. For clarity, the following sections focus on three core combat variables: health points, skill meters, and attack strengths.

\subsection{Agentic Control Layer}
\label{subsec:gamedirector_control}

The agentic control layer bridges player-configured mechanics and generated gameplay through visual understanding, state tracking, action planning, and rule enforcement. It integrates visual feedback and player inputs with explicit combat rules, while intelligently controlling the boss to produce structured action prompts for the world model.

\noindent\textbf{Visual Understanding.}
To ground decisions in visual observations, two lightweight ResNet-18 models, namely SituationNet $\mathcal{S}_{\phi}$ and AttackNet $\mathcal{A}_{\phi}$, are integrated to analyze the combat context and detect attack signals from the generated rollout. Both networks take the three most recent gameplay frames as input, but operate at different frequencies:
\begin{equation}
\begin{aligned}
    (\hat{d}_t,\hat{\theta}_t)
    &=
    \mathcal{S}_{\phi}(\mathbf{F}_{t-2:t}),
    && t\in\{80n\}_{n\geq 1}, \\
    (\hat{a}_t^p,\hat{a}_t^b)
    &=
    \mathcal{A}_{\phi}(\mathbf{F}_{t-2:t}),
    && t\in\{4n\}_{n\geq 1}.
\end{aligned}
\end{equation}
Specifically, SituationNet $\mathcal{S}_{\phi}$ estimates the distance $\hat{d}_t$ and relative facing direction $\hat{\theta}_t$ between the boss and player. These estimates are treated as spatial states for the LLM to select the most suitable skill. AttackNet $\mathcal{A}_{\phi}$ predicts $\hat{a}_t^p,\hat{a}_t^b\in\{0,1\}$, indicating whether the player or boss is hit by its counterpart, and provides evidence for numerical state updates. $\mathcal{S}_{\phi}$ processes the three most recent frames every 80 frames (i.e., 5s), whereas $\mathcal{A}_{\phi}$ processes them every 4 frames (i.e., 0.25s). This much higher execution frequency enables $\mathcal{A}_{\phi}$ to capture short-lived attack events that require finer-grained temporal detection than changes in the overall combat situation.

\begin{figure*}[t]
\centering
    \includegraphics[width=\textwidth]{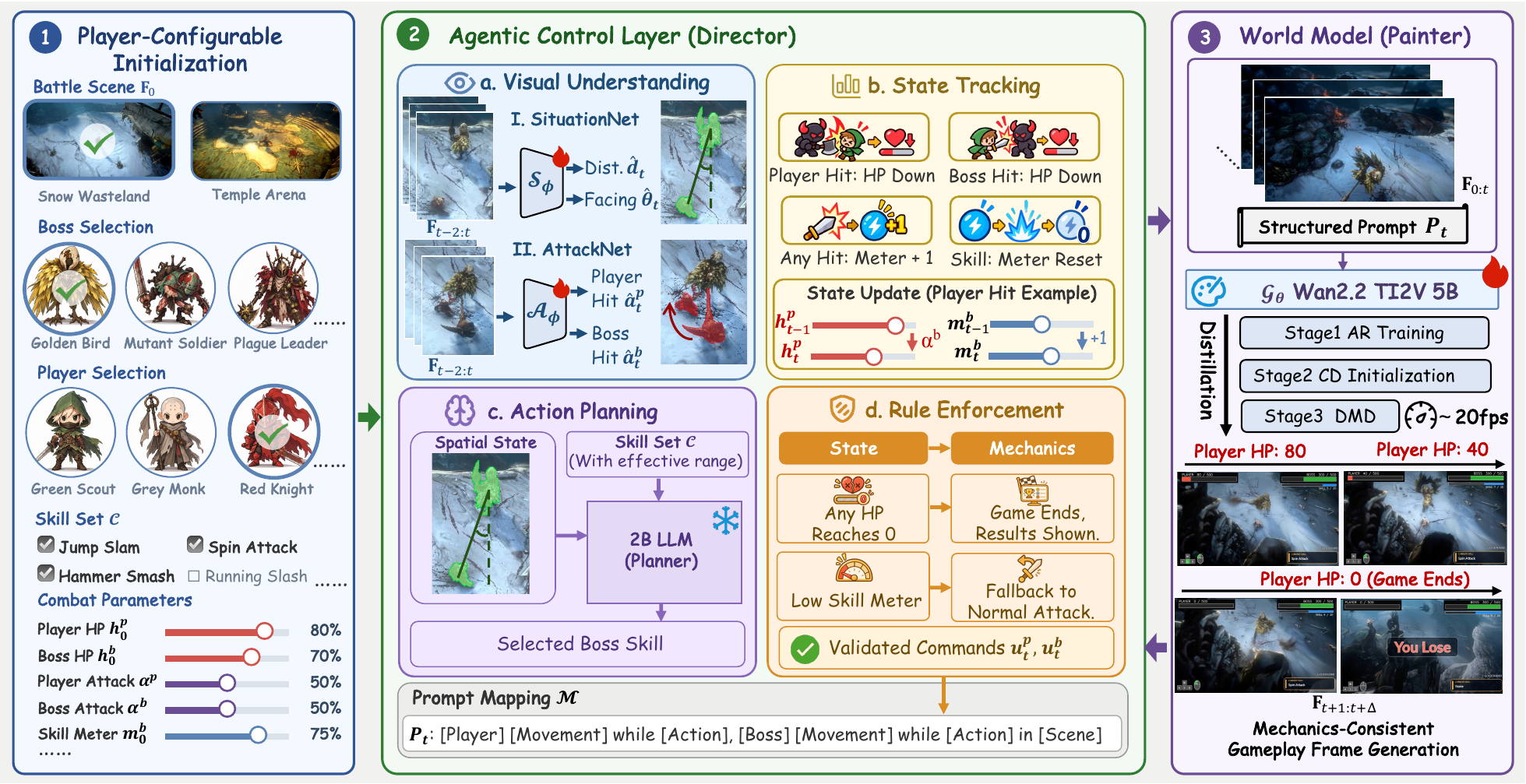}
    \caption{
    Framework of \textbf{GameDirector}. Player-defined configurations are processed by an agentic control layer that performs visual understanding, state tracking, action planning, and rule enforcement, producing structured prompts for mechanics-consistent world-model generation.
    }
    \label{fig:method}
\end{figure*}

\noindent\textbf{State Tracking.}
Unlike world models trained solely with pixel-space supervision, which may overlook explicit state progression, the state tracker uses the attack signals $\hat{a}_t^p,\hat{a}_t^b\in\{0,1\}$ to update the health points $h_t^p,h_t^b$ and the boss skill meter $m_t^b$. Specifically, the health points are updated as
\begin{equation}
    h_{t+1}^p
    =
    \max\left(0,h_t^p-\alpha^b\hat{a}_t^p\right),
    \qquad
    h_{t+1}^b
    =
    \max\left(0,h_t^b-\alpha^p\hat{a}_t^b\right),
    \label{eq:health_update}
\end{equation}

where $\alpha^p$ and $\alpha^b$ denote the configured attack strengths of the player and boss, respectively. The boss skill meter increases when the boss either hits the player or is hit by the player:
\begin{equation}
    m_{t+1}^b
    =
    \min\left(M,m_t^b+\hat{a}_t^p+\hat{a}_t^b\right),
    \label{eq:skill_meter_update}
\end{equation}
where $M$ denotes the maximum skill-meter value. Thus, the configured attack strength determines the health deduction whenever a hit is detected, while either type of successful hit increases the boss skill meter. These mechanics follow common conventions in combat games rather than being tailored to a specific title.

\noindent\textbf{Action Planning.}
Instead of overfitting to the boss behavior in the training set, boss should be able to perform skills based on the context. To strategically control the boss, a 2B language model is used as a planner  to select skills according to the spatial context and the player-configured skill set $\mathcal{C}$. The planner uses the latest distance $\hat{d}_t$ and relative facing direction $\hat{\theta}_t$ from SituationNet $\mathcal{S}_{\phi}$, together with the skill descriptions in $\mathcal{C}$, to make a decision:
\begin{equation}
    c_t^b=LLM\left(\hat{d}_t,\hat{\theta}_t,\mathcal{C}\right),
    \label{eq:action_planning}
\end{equation}
where $c_t^b\in\mathcal{C}$ is the selected boss skill by the language model. Skill descriptions and their effective ranges allow the language model to select an appropriate skill based on the player’s position and the boss’s facing direction. For example, the boss may use a gap-closing skill when the player is far away or a melee attack when the player is within range. The selected skill is then passed to rule enforcement for validation and execution. Notably, leveraging a 2B language model for strategic boss control provides a flexible interface that can be extended to incorporate richer state representations and support more fine-grained, context-aware action planning.

\noindent\textbf{Rule Enforcement.}
After receiving the player input and planning the boss action, the rule enforcement module performs a final validation of both characters' actions and movements. When either health point $h_t^p$ or $h_t^b$ reaches zero, the death rules override all ongoing actions and player inputs: the defeated character remains in the death state. In addition, the boss can perform a special skill only when its current skill meter $m_t^b$ reaches the corresponding activation threshold $\tau$ and the skill's cooldown has elapsed; otherwise, it falls back to a normal attack. These constraints produce the final validated action--movement commands $u_t^p$ and $u_t^b$, ensuring that the generation prompts consistently comply with the game mechanics.
\vspace{-5pt}
\subsection{Controllable Generation}
\label{subsec:gamedirector_generation}

The validated player and boss action--movement pairs $u_t^p$ and $u_t^b$ are mapped to a natural-language prompt $P_t$, which conditions a DiT-based world model to generate the next video segment:
\begin{equation}
    P_t=\mathcal{M}(u_t^p,u_t^b),
    \qquad
    \mathbf{F}_{t+1:t+\Delta}
    =\mathcal{G}_{\theta}
    (\mathbf{F}_{0:t},P_t,\boldsymbol{\epsilon}_t).
    \label{eq:controllable_generation}
\end{equation}
Here, $\mathcal{M}$ maps actions to the template-based prompts, $\mathcal{G}_{\theta}$ denotes the video generator, $\boldsymbol{\epsilon}_t$ is Gaussian noise, and $\Delta=4$ is the number of frames generated per step. The generated frames are fed back to the agentic control layer for state updates and subsequent decisions, closing the loop between game mechanics and visual generation.

Overall, our proposed framework, GameDirector, preserves player configurability while enforcing essential game rules. It continuously monitors the combat situation, reliably updates key states, and adaptively controls the boss, guiding the world model to generate frames that balance creative flexibility with mechanical consistency.


\vspace{-5pt}
\section{Experiments}


\vspace{-5pt}
\subsection{Implementation Details}
We initialize the video world model from Wan2.2-TI2V-5B \citep{wan2025wan} and fine-tune its DiT across three games for 60K steps, while freezing the VAE and text encoder \citep{chung2023unimax}. The model takes the first frame and 20 cell-level prompts to generate 80 frames at $832\times480$. We train with AdamW at $5\times10^{-5}$ on four GPUs, followed by three-stage Causal Forcing++ distillation \citep{zhao2026causal}. AttackNet and SituationNet use ResNet-18 \citep{he2016deep} with a single-layer GRU and are jointly trained across all three games for 20 epochs. Action planning uses Gemma-4-E2B-it \citep{team2026gemma} without task-specific fine-tuning. A single shared checkpoint is used for each module across all games, enabling real-time closed-loop gameplay at 20 FPS. Please refer to Appendix \ref{app:implementation} for more agentic implementation details.

\vspace{-5pt}
\subsection{Evaluation of GameDirector}
We evaluate GameDirector through three questions. \textbf{Q1} (Tab. \ref{tab:main_results}): Does the agentic framework improve state tracking, mechanics fidelity, and boss control compared with alternative game world model designs? \textbf{Q2} (Tab. \ref{tab:visual_understanding}): Can the visual understanding module reliably capture gameplay context and support agentic control across different games? \textbf{Q3} (Sec. \ref{subsec:visual}): Are the benefits of the agentic framework perceptible to players during actual gameplay?

\begin{table*}[t]
\centering
\caption{
Comparison of different state modeling and boss control strategies. Each setting adopts a representative method with minimal adaptation, and all methods are trained using the same dataset. We use $N=3$ VLM evaluations and ``--'' denotes metrics that are not applicable to a given method.
}
\label{tab:main_results}

\setlength{\tabcolsep}{4pt}
\renewcommand{\arraystretch}{1.05}

\resizebox{\textwidth}{!}{
\begin{tabular}{lcc|cc|c|cc|cc}
\toprule
& \multicolumn{2}{c|}{\textbf{State Modeling}}
& \multicolumn{2}{c|}{\textbf{Boss Control}}
& \multirow{2}{*}{\textbf{State Align. (\%) $\uparrow$}}
& \multicolumn{2}{c|}{\textbf{Mechanics Fidelity (\%) $\uparrow$}}
& \multicolumn{2}{c}{\textbf{Decision Quality (\%) $\uparrow$}} \\
\cmidrule(lr){2-3}
\cmidrule(lr){4-5}
\cmidrule(lr){7-8}
\cmidrule(lr){9-10}

\textbf{Setting}
& \textbf{None}
& \textbf{Predictive}
& \textbf{Implicit}
& \textbf{Explicit}
&
& \textbf{GPT-5.5}
& \textbf{Gemini}
& \textbf{GPT-5.5}
& \textbf{Gemini} \\

\midrule
(1) \citep{he2025matrix}
& $\checkmark$
& 
& $\checkmark$
& 
& --
& $25.1_{\pm 0.7}$
& $25.0_{\pm 0.3}$
& $28.1_{\pm 0.9}$
& $26.4_{\pm 0.2}$ \\

(2) \citep{wang2026reactivegwm}
& $\checkmark$
& 
& 
& $\checkmark$
& --
& $21.4_{\pm 0.2}$
& $21.4_{\pm 0.1}$
& $48.5_{\pm 0.9}$
& $47.2_{\pm 0.3}$\\

(3) \citep{li2026wildworld}
& 
& $\checkmark$
& $\checkmark$
& 
& $70.38$
& $34.3_{\pm 0.5}$
& $34.1_{\pm 0.4}$
& $27.1_{\pm 0.4}$
& $26.6_{\pm 0.3}$ \\

(4) \citep{lin2026stateplay}
& 
& $\checkmark$
& 
& $\checkmark$
& $72.01$
& $28.1_{\pm 0.3}$
& $28.2_{\pm 0.1}$
& $44.3_{\pm 1.0}$
& $44.6_{\pm 0.1}$\\

\midrule

\textbf{GameDirector} (Ours)
& \multicolumn{2}{c|}{\textbf{Agentic}}
& \multicolumn{2}{c|}{\textbf{Agentic}}
& $\mathbf{100.0}$
& $\mathbf{98.5_{\pm 0.2}}$
& $\mathbf{99.6_{\pm 0.2}}$
& $\mathbf{88.4_{\pm 0.7}}$
& $\mathbf{93.0_{\pm 0.6}}$ \\

\bottomrule
\end{tabular}
}
\end{table*}

\noindent\textbf{Evaluation Metrics.}
We evaluate GameDirector from four complementary aspects: state alignment, mechanics fidelity, decision quality, and visual understanding.

\begin{itemize}[leftmargin=*]

\item \textbf{State Alignment:} Measures whether predicted game states follow the underlying gameplay mechanics, including whether HP decreases following valid hits and whether the skill meter increases or resets under the corresponding events. Since generated rollouts do not provide ground-truth hit annotations, we use AttackNet, which achieves over 90\% accuracy on held-out data, to infer hit events as pseudo ground truth for evaluating state transitions.

\item \textbf{Mechanics Fidelity:} Measures whether generated gameplay follows necessary rules, including termination behavior as HP reaches 0 and skill cooldown rules. GPT-5.5 \citep{openai2026gpt55} and Gemini-3.1-Pro \citep{google2026gemini31pro} are used as VLM judges. The ground-truth termination image for each player and boss is provided to the VLM as visual context.

\item \textbf{Decision Quality:} Measures whether the boss selects skills appropriate for the current spatial context and available skill set, as judged by GPT-5.5 and Gemini-3.1-Pro. A description of each skill, including its effective range, is provided as context. 


\item \textbf{Visual Understanding:} Evaluates the reliability of the perception modules in the agentic control layer. AttackNet measures hit detection performance, while SituationNet measures spatial understanding, including relative distance, angle, actionable range, and direction.

\end{itemize}

The first three metrics are reported in Tab.~\ref{tab:main_results}, while visual understanding results are reported in Tab.~\ref{tab:visual_understanding}. Detailed evaluation protocols are provided in the Appendix \ref{app:evaluation}.

\noindent\textbf{Effectiveness of Agentic Control.}
As shown in Tab.~\ref{tab:main_results}, we compare our agentic framework with alternative state modeling and boss control strategies commonly used in existing game world models. Prior methods either overlook explicit states or predict them directly, while boss behavior is either learned implicitly from training data or guided by high-level commands explicitly. Details of baseline methods implementation are provided in the Appendix \ref{app:baseline_details}.


For \textbf{state alignment}, predictive state modeling performs reasonably well on short clips but degrades over 1-minute rollouts, reaching only about 70\% in Settings (3) and (4). In contrast, our agentic framework deterministically updates states from AttackNet-detected hit events, ensuring exact consistency with these detected signals throughout the rollout. Thus, the reported 100\% reflects deterministic state-update consistency, while errors in the underlying hit detection are separately characterized by AttackNet accuracy. Even accounting for such perception errors, the agentic approach remains more reliable than predictive state modeling. A similar trend appears in \textbf{mechanics fidelity}, where predictive state modeling improves by less than 10\% over pure pixel-space fitting during long-horizon rollouts. In contrast, GameDirector decouples state tracking and rule enforcement from visual generation, achieving over 98\% mechanics fidelity under both VLM judges.

For \textbf{decision quality}, implicit boss control in Settings (1) and (3) mainly reproduces behaviors from the training data, while explicit control in Settings (2) and (4) improves performance through strategy-aware natural-language commands. GameDirector further improves decision quality by more than 39.9\%, reaching 88.4\% and 93.0\% accuracy under the two VLM judges. This shows that agentic reasoning over the current gameplay situation enables boss actions to better match the spatial context than either implicit behavior modeling or explicit language-based control. Additional evaluation results, deployment settings, and inference speed analysis are provided in Appendix~\ref{app:exp_results}.

\begin{table*}[t]
\centering
\caption{
Performance of AttackNet and SituationNet in Visual Understanding across three games.
}
\label{tab:visual_understanding}

\setlength{\tabcolsep}{5pt}
\renewcommand{\arraystretch}{1.05}

\resizebox{\textwidth}{!}{
\begin{tabular}{lcc|ccccc}
\toprule
& \multicolumn{2}{c|}{\textbf{AttackNet}}
& \multicolumn{5}{c}{\textbf{SituationNet}} \\
\cmidrule(lr){2-3}
\cmidrule(lr){4-8}

\textbf{Game}
& \textbf{Macro-F1 $\uparrow$}
& \textbf{Accuracy (\%) $\uparrow$}
& \textbf{Dist. MAE $\downarrow$}
& \textbf{Angle MAE ($^\circ$) $\downarrow$}
& \textbf{Front-Cone (\%) $\uparrow$}
& \textbf{Decision-Zone (\%) $\uparrow$}
& \textbf{Tactical-Sector (\%) $\uparrow$} \\
\midrule

Game N
& $0.8476$
& $90.53$
& $0.3556$
& $10.91$
& $89.29$
& $87.06$
& $86.95$ \\

Game V
& $0.9135$
& $92.48$
& $0.1876$
& $2.22$
& $96.72$
& $96.31$
& $96.31$ \\

Game H
& $0.8871$
& $96.11$
& $0.2637$
& $6.39$
& $96.49$
& $96.49$
& $96.49$ \\

\midrule
\textbf{Average}
& $\mathbf{0.8906}$
& $\mathbf{92.94}$
& $\mathbf{0.2690}$
& $\mathbf{6.51}$
& $\mathbf{94.17}$
& $\mathbf{93.29}$
& $\mathbf{93.25}$ \\

\bottomrule
\end{tabular}
}
\end{table*}

\noindent\textbf{Reliability of Visual Understanding.}
We evaluate AttackNet and SituationNet, which provide the attack signal and spatial states used by subsequent control modules. As shown in Table~\ref{tab:visual_understanding}, AttackNet achieves an average Macro-F1 of 0.8906 and an accuracy of 0.9294. SituationNet obtains an average distance MAE of 0.2690 and angle MAE of 6.51$^\circ$, while achieving over 93\% accuracy on all spatial classification metrics: Front-Cone identifies whether the target is ahead, Decision-Zone determines whether it is within the actionable range, and Tactical-Sector classifies its relative direction. The consistent performance across all three games demonstrates that the shared agent architecture reliably supports state tracking and action planning.

\label{subsec:visual}
\begin{figure*}[t]
\centering
    \includegraphics[width=\textwidth]{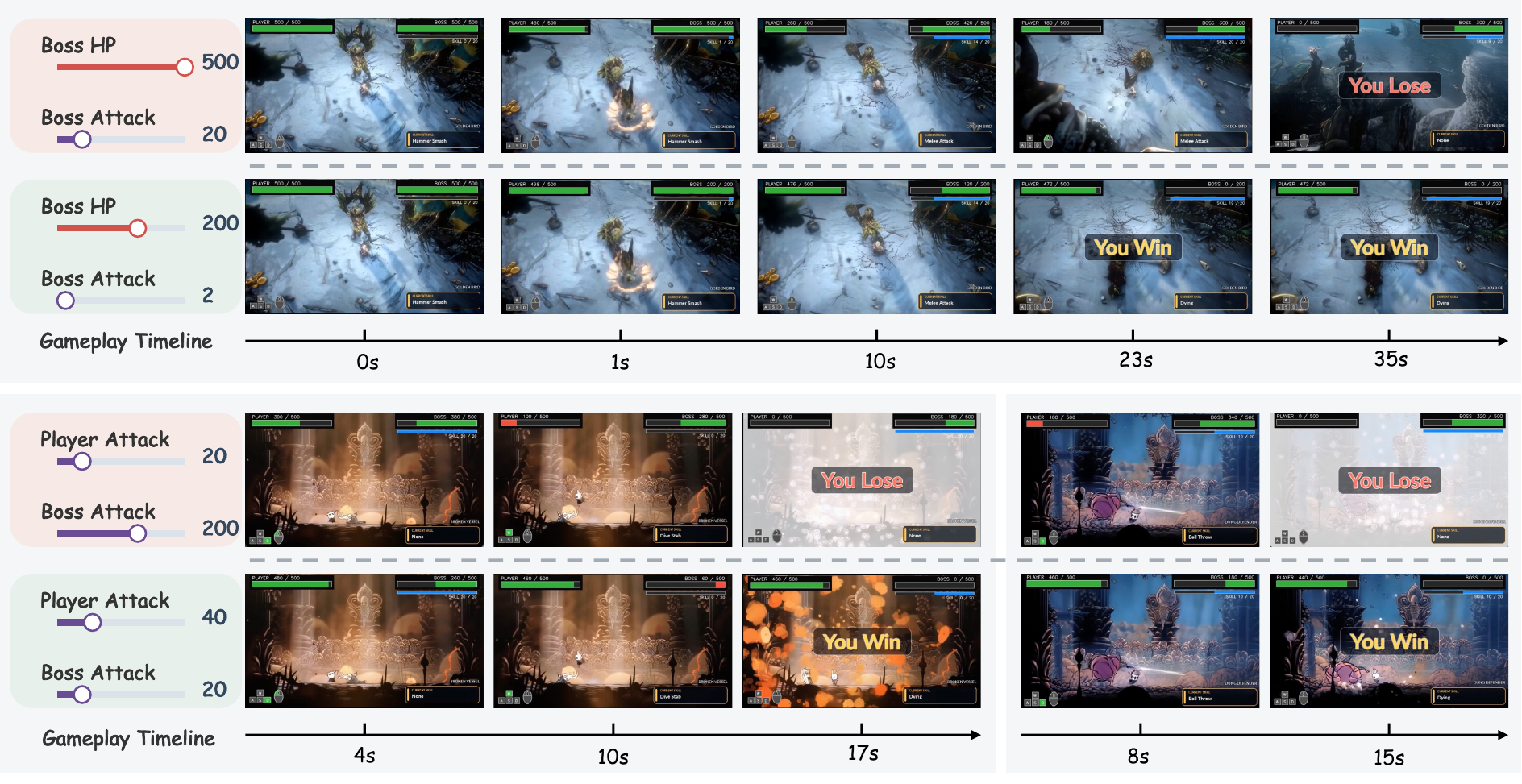}
    \caption{
    Comparison of gameplay trajectories under different player-defined state initializations. The UI overlays are added only for visualization and dynamically reflect the explicit states tracked by the agentic control layer. Please zoom in for better visibility.
    }
    \label{fig:visual}
\end{figure*}
\vspace{-10pt}
\subsection{Qualitative Visualization}

\noindent\textbf{Player-Configurable Initialization.}
To evaluate whether player-defined state configurations are faithfully reflected in gameplay, Fig.~\ref{fig:visual} compares gameplay trajectories under different initial combat settings. The world model itself generates only the visual frames. For clarity, we overlay the health bars, skill meters, and action labels, which are directly derived from the explicit states maintained by the agentic control layer.

Across three examples, the visual gameplay evolution remains consistent with the tracked states: health decreases after successful attacks, while skill meters increase as the interaction progresses. More importantly, changing the player-defined initialization leads to corresponding changes in both state evolution and gameplay outcome. In the top example, reducing the boss HP and attack strength changes the outcome from a player loss to a win. In the bottom two examples, increasing the player's attack strength while reducing the boss's attack strength similarly reverses the combat outcome. These results demonstrate that GameDirector can reliably translate player-configured combat parameters into consistent state transitions and observable gameplay consequences. Please refer to Appendix \ref{app:subsec_more_visual} for more visualization results and our anonymous project page for video demos.

\begin{figure*}[t]
\centering
    \includegraphics[width=\textwidth]{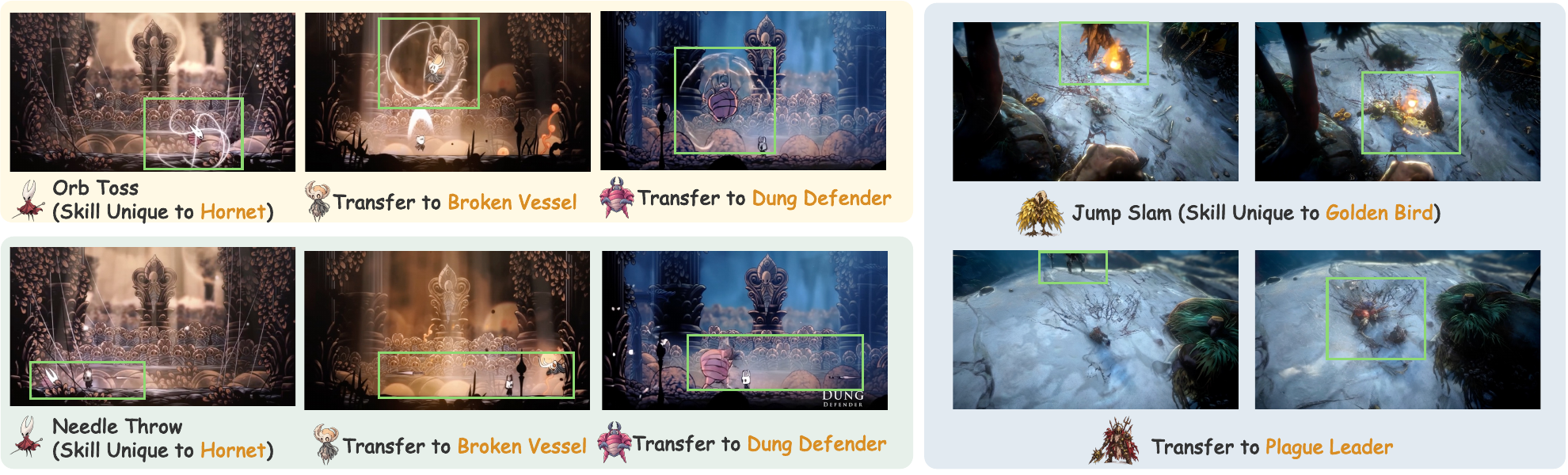}
    \caption{
    Cross-character skill transfer with GameDirector. Skills originally associated with one boss can be assigned to different bosses through player-configurable initialization.
    }
    \label{fig:skill}
\end{figure*}

\noindent\textbf{Cross-Character Skill Transfer.}
The agentic design of GameDirector allows players to assign skills from one boss to another during player-configurable initialization, including skills that are not originally available to the selected boss. As shown in Fig.~\ref{fig:skill}, GameDirector can execute these transferred skills across different characters during gameplay, enabling flexible boss--skill customization and more diverse gameplay experiences.

\begin{figure*}[t]
\centering
    \includegraphics[width=\textwidth]{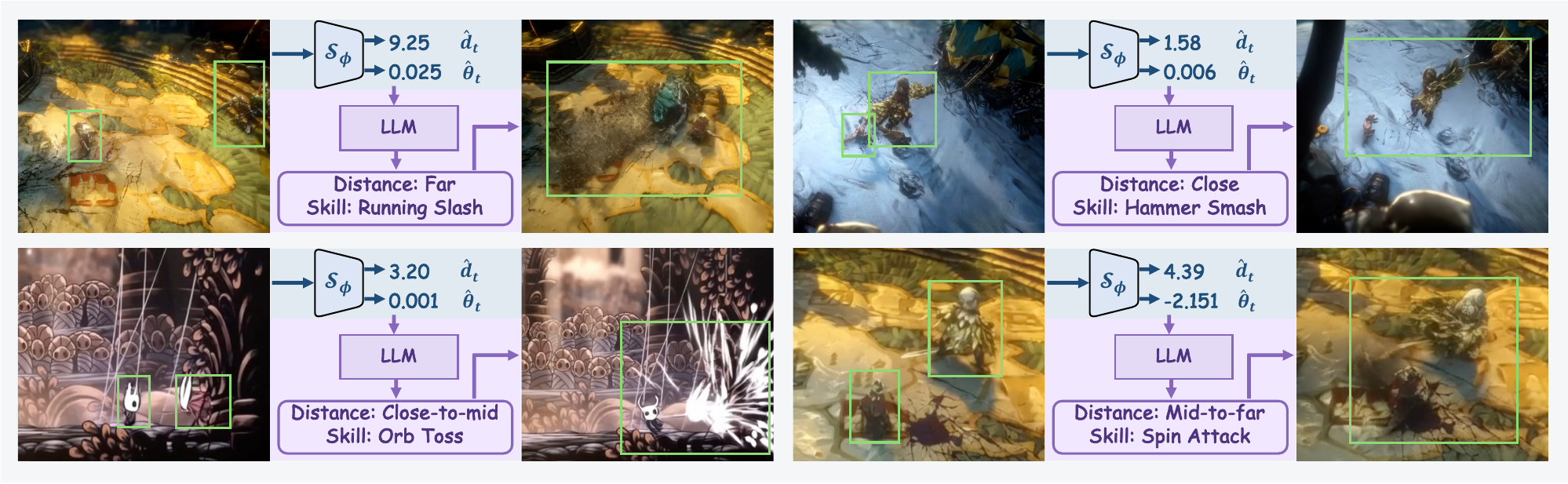}
    \caption{
    Adaptive boss action planning. GameDirector selects boss skills based on the spatial relation between the player and boss, adapting actions across different distances.
    }
    \label{fig:control}
\vspace{-10pt}
\end{figure*}
\noindent\textbf{Adaptive Boss Action Planning.} 
Fig. \ref{fig:control} illustrates how visual understanding and action planning in the agentic control layer work together to determine the boss action tactically based on the current spatial states. By considering the distance between the player and the boss, the language model selects an appropriate skill, resulting in more context-aware boss behavior that better aligns with how boss actions are designed in real games.

\vspace{-10pt}
\section{Conclusions and Discussions}
\vspace{-8pt}
In this paper, we introduce \textbf{GameDirector}, an agentic framework that decouples hard-coded game logic from end-to-end game world models, combining the reliability of traditional engines with the flexibility of generative modeling. Its agentic control layer provides accurate visual understanding, consistent state tracking, context-aware boss control and reliable rule enforcement. These capabilities transfer across three games and support player-configurable combat initialization. Experiments show significant gains over representative world model settings with different state modeling and boss control strategies, highlighting the value of explicit agentic control for coherent gameplay.

Concurrent works use game engines or coding agents to build structured intermediate representations, such as bounding boxes \citep{chen2026code, huang2026programmable}, articulated skeletons \citep{meng2026marionette}, or coarse 3D geometry \citep{zhan2026magpie}, for more stable generation. In future work, we aim to extend GameDirector to such structured and controllable settings, allowing the agentic framework to further empower player-defined game generation with greater flexibility, reliability, and mechanics consistency.



\bibliography{iclr2027_conference}
\bibliographystyle{iclr2027_conference}

\appendix
\section{Appendix}
\subsection{Dataset Details}
\label{app:dataset}

\setcounter{table}{4}
\begin{table}[h]
\centering
\small
\caption{Dataset statistics. Percentages are computed over all 51,786 clips,
and hours are based on five-second clips.}
\label{tab:dataset_manifest_stats}
\begin{tabular}{lrrrrr}
\toprule
Game & Train & Eval. & Total & Share & Hours \\
\midrule
Game N & 22,605 & 100 & 22,705 & 43.8\% & 31.53 \\
Game V & 18,557 & 100 & 18,657 & 36.0\% & 25.91 \\
Game H & 10,324 & 100 & 10,424 & 20.1\% & 14.48 \\
\midrule
Total & 51,486 & 300 & 51,786 & 100.0\% & 71.93 \\
\bottomrule
\end{tabular}
\end{table}

\begin{table}[h]
\centering
\small
\caption{Per-character and per-boss coverage in the dataset.}
\label{tab:entity_hours}
\resizebox{\textwidth}{!}{%
\begin{tabular}{lllrr@{\qquad}lllrr}
\toprule
Role & Game & Entity & Clips & Hours & Role & Game & Entity & Clips & Hours \\
\midrule
Player & N & Green Scout & 5,835 & 8.10 & Boss & N & Golden Bird & 10,642 & 14.78 \\
Player & N & Grey Monk & 11,284 & 15.67 & Boss & N & Mutant Soldier & 7,148 & 9.93 \\
Player & N & Red Knight & 5,586 & 7.76 & Boss & N & Plague Leader & 4,915 & 6.83 \\
Player & V & Vampire Lord & 18,657 & 25.91 & Boss & V & Harpy Queen & 18,657 & 25.91 \\
Player & H & the Knight & 10,424 & 14.48 & Boss & H & Hornet & 3,465 & 4.81 \\
&&&&& Boss & H & Broken Vessel & 3,308 & 4.59 \\
&&&&& Boss & H & Gruz Mother & 1,470 & 2.04 \\
&&&&& Boss & H & Dung Defender & 2,181 & 3.03 \\
\midrule
\multicolumn{3}{r}{Player-side total} & 51,786 & 71.93 &
\multicolumn{3}{r}{Boss-side total} & 51,786 & 71.93 \\
\bottomrule
\end{tabular}
}
\end{table}

\noindent\textbf{Dataset Collection.} We collect synchronized gameplay data from Game N (\textit{No Rest for the Wicked}), Game V (\textit{Vampire}\footnote{Due to copyright considerations, we anonymize this game throughout the paper. Its original title and visual examples are not disclosed; only experimental results and statistics are reported.}), and Game H (\textit{Hollow Knight}) using the automated
agent described in Sec.~\ref{subsec:dataset}. The agent controls the playable
character with keyboard and mouse inputs and repeats each encounter across
different characters, bosses, and scenes. Alongside the rendered video, we
record health points, skill meters, executed actions, character positions, and
facing directions. All signals are timestamped and aligned to the video
timeline. Game-specific action identifiers are mapped to a fixed action
vocabulary, and clips with incomplete states or unknown actions are removed.

Each sample is a non-overlapping five-second clip with 80 frames at 16 FPS and
resolution \(832\times480\). We divide the clip into 20 intervals of 0.25 s.
Continuous states are interpolated to this time grid, while discrete actions
are assigned to the interval in which they occur. Each interval contains the
player and boss actions, movement, health points, spatial states, control
inputs, and the text prompt used to train the world model. The last incomplete
clip of a fight is padded with its final valid frame and state, without using
content from the next encounter.

Tab.~\ref{tab:dataset_manifest_stats} summarizes the final dataset. It contains
51,786 clips, including 51,486 training clips and 300 evaluation clips. The
evaluation set contains 100 clips per game and does not overlap with the
training manifests. For Game N, the evaluation split covers all
nine playable-character--boss combinations in both scenes. The Game V and Game
H evaluations use held-out clips from their corresponding sessions.

Tab.~\ref{tab:entity_hours} further reports the coverage of each character and
boss. A clip is counted once for its playable character and once for its boss,
so the player-side and boss-side totals are both equal to the complete dataset.

\noindent\textbf{Skill Set}. For action planning, we construct one skill card for each special skill. The
card describes its role, duration, cooldown, effective range, movement effect,
and meter cost. These values are estimated only from the aligned training
clips. Tab.~\ref{tab:skill_cards} lists the main numeric fields. Game V uses
melee and repositioning actions because its retained data does not contain a
named special skill.

\begin{table}[t]
\centering
\small
\caption{Special-skill cards used for action planning.}
\label{tab:skill_cards}
\resizebox{\textwidth}{!}{%
\begin{tabular}{llllcc}
\toprule
Game & Boss & Special Skill & Role & Duration / cooldown (s) & Best range \\
\midrule
N & Golden Bird & Hammer Smash & area punish & 7.00 / 14.50 & 1.1--5.5 \\
N & Golden Bird & Jump Slam & short engage & 4.50 / 18.25 & 0.0--3.0 \\
N & Golden Bird & Spin Attack & area denial & 9.25 / 20.00 & 2.5--7.0 \\
N & Mutant Soldier & Back Spin & rear punish & 3.75 / 5.50 & 1.0--4.0 \\
N & Mutant Soldier & Front Spin & frontal sweep & 4.25 / 4.75 & 2.0--7.0 \\
N & Mutant Soldier & Running Slash & far engage & 1.50 / 5.25 & 5.5--9.0 \\
N & Plague Leader & Left Spin & left punish & 3.00 / 3.25 & 1.0--3.5 \\
N & Plague Leader & Right Spin & right punish & 3.00 / 3.00 & 1.0--4.0 \\
N & Plague Leader & Shield Charge & far engage & 5.50 / 16.25 & 3.5--8.5 \\
\midrule
H & Hornet & Aerial Dash & short engage & 0.75 / 1.75 & 0.5--3.6 \\
H & Hornet & Needle Throw & mid ranged & 2.50 / 4.25 & 2.8--5.1 \\
H & Hornet & Orb Toss & mid projectile & 1.75 / 2.50 & 2.0--4.1 \\
H & Broken Vessel & Dive Stab & short engage & 1.75 / 2.50 & 0.1--3.3 \\
H & Broken Vessel & Overhead Slash & mid punish & 2.50 / 3.00 & 2.7--5.5 \\
H & Gruz Mother & Shoulder Charge & gap close & 1.75 / 3.25 & 0.4--5.0 \\
H & Dung Defender & Ground Burst & area punish & 1.75 / 4.75 & 1.3--5.2 \\
H & Dung Defender & Rolling Dive & mobile pressure & 3.00 / 0.75 & 1.3--6.2 \\
\bottomrule
\end{tabular}
}
\end{table}

\noindent\textbf{Structured Text Prompt.} We convert game-specific raw action identifiers into a shared canonical vocabulary using per-game mapping tables.
  Player actions are derived from keyboard and mouse inputs, while boss actions are mapped from logged attack or skill
  events. Position changes are discretized into directional movement labels. These attributes are then instantiated in
  a unified structured template describing each character’s identity, movement, and action, for example, “[player]
  moves left while performing melee attack, [boss] stays stationary while performing hammer smash.” Health states
  additionally determine terminal prompts: once a character’s health reaches zero, its action is overridden with
  “dying” and its movement is set to stationary, while the surviving character is assigned a non-attacking prompt.
  This rule ensures that the textual conditions remain consistent with the recorded combat outcome.
\subsection{Implementation Details}
\label{app:implementation}

\noindent\textbf{Visual understanding.}
AttackNet and SituationNet use an ImageNet-pretrained ResNet-18 followed by a
unidirectional GRU with hidden size 384. Both models take three consecutive
frames resized to \(336\times192\). They are trained jointly across the three
games for 20 epochs using AdamW, a learning rate of \(3\times10^{-4}\), weight
decay \(10^{-4}\), and dropout 0.1.

AttackNet predicts whether the player or boss is hit in each 0.25-s interval.
The labels are derived from aligned health-point changes, and weighted binary
cross entropy is used to address class imbalance. SituationNet predicts
normalized distance and the sine and cosine of the relative angle in each 5-s interval. Invalid and
post-death intervals are excluded from its training. The same checkpoint of
each model is used for all three games.

\noindent\textbf{Action planning.}
We use Gemma-4-E2B-it~\citep{team2026gemma} as the planner with greedy decoding
and no task-specific fine-tuning. At each decision point, it receives the
latest spatial estimate, combat state, action history, and the available skill
cards. Its output must select one action from the provided menu. The selected
action is then checked against the current meter and cooldown before being
passed to the generative model.

\noindent\textbf{Rule Following.} The external state tracker stores player health, boss health, attack strengths,
and the boss skill meter. AttackNet signals trigger deterministic health and
meter updates. A special skill requires the corresponding skill-meter threshold to be reached and a valid
cooldown, and its release consumes the meter. Once either character reaches
zero health, new attacks and state changes are disabled and the death action
overrides subsequent prompts.

\noindent\textbf{Video World model.}
We initialize the video world model from Wan2.2-TI2V-5B~\citep{wan2025wan}. The VAE and
text encoder are frozen, while all DiT parameters are trained on the three-game
dataset for 60K steps using AdamW and a learning rate of
\(5\times10^{-5}\). Each training clip contains one clean conditioning frame
and 20 predicted latent frames. The corresponding 20 action prompts are
encoded separately and aligned with these latent frames through local
cross-attention. This alignment allows the model to render actions at the
0.25-s interval used by the control layer.

For causal generation, we follow the three-stage Causal Forcing++ procedure
\citep{zhao2026causal}. We first train an autoregressive teacher for 20K steps,
then apply consistency distillation for 12K steps, followed by DMD training for
5K steps. The final generator uses three denoising steps and produces four frames
per control interval.

\begin{figure}[t]
      \centering
\begin{tcolorbox}[
      width=\linewidth,
      colback=white,
      colframe=gray!35,
      colbacktitle=gray!25,
      coltitle=black,
      title=\textbf{Evaluation Prompt (Mechanics Fidelity)},
      fonttitle=\normalsize\bfseries,
      boxrule=0.8pt,
      arc=3mm,
      left=1mm,
      right=1mm,
      top=0.8mm,
      bottom=0.8mm
  ]
  You are evaluating mechanics consistency in a generated boss-fight rollout.

  The provided clip begins shortly before AttackNet determines that one
  character has crossed the death threshold and continues until the end of
  the rollout.

  \textbf{Fight:} [player] vs.\ [boss] in [location].\\
  \textbf{Dying character:} [character].\\
  \textbf{Death-threshold time:} [timestamp].

  Determine whether the dying character enters and maintains a valid death
  state after the threshold. A valid state includes either:

  \begin{itemize}[
      leftmargin=5mm,
      itemsep=0pt,
      topsep=1pt,
      parsep=0pt
  ]
      \item remaining collapsed or fallen; or
      \item becoming static and no longer participating in combat.
  \end{itemize}

  If the character continues walking, attacking, dodging, or actively
  fighting for a meaningful portion of the remaining clip, the death state
  is inconsistent.

  A reference image of the character's death pose is attached as visual
  guidance, but an exact pose match is not required.

  \textbf{Return one JSON object only.} Do not include markdown or additional
  text.

  \end{tcolorbox}
      \caption{VLM prompt for rollout Mechanics Fidelity evaluation.}
      \label{fig:rollout_mechanics_prompt}
  \end{figure}
\subsection{Evaluation Details}
\label{app:evaluation}

\noindent\textbf{Closed-loop rollouts.}
We evaluate long-horizon gameplay on 100 held-out Game-N initial conditions,
covering three playable characters, three bosses, and two scenes. All methods
receive the same initial frame and one-minute player-control sequence. A
rollout contains at most twelve five-second segments, and each new segment
starts from the final generated frame of the preceding segment. GameDirector
uses initial player and boss health of 500, attack strengths of 40 and 20, and
a full initial boss meter. A rollout stops after a character dies and a final
five-second death segment is generated.

\noindent\textbf{State Alignment.}
State alignment measures whether explicit state transitions are consistent with the gameplay signals observed from generated frames. Each 60-s rollout is evaluated every 0.25 s, yielding 240 ticks. At each tick, AttackNet predicts player and boss hit events from the latest three frames, and we verify three constraints: player HP decreases only after a detected player hit, boss HP decreases only after a detected boss hit, and the boss skill meter increases upon successful hits and resets after skill execution. Accuracy is averaged over all rollouts, ticks, and state variables. Because GameDirector deterministically updates states using the same AttackNet signals used for evaluation, its perfect alignment score reflects logical consistency between perception and state updates rather than perfect recovery of hidden engine states.

\noindent\textbf{Mechanics Fidelity.}
Mechanics fidelity measures whether discrete game rules remain satisfied throughout long-horizon rollouts. We evaluate two complementary aspects: skill legality and death persistence. A GameDirector skill release is valid only when the required skill meter is available and the corresponding cooldown has elapsed. For baselines without explicit skill labels, an evaluation-only ResNet-18 SkillClassifier identifies visible boss skills from generated frames with over 95\% accuracy on held-out data, allowing us to verify that consecutive special skills respect the minimum cooldown interval. Death persistence requires a character whose HP reaches zero to subsequently enter a death state and remain non-attacking. GPT-5.5 and Gemini-3.1-Pro independently judge the rendered death intervals, and the results are reported over three runs. The prompt is shown in Fig. \ref{fig:rollout_mechanics_prompt}.

\noindent\textbf{Decision Quality.}
Decision quality evaluates whether the boss selects a skill appropriate for the current combat situation. SituationNet provides the relative distance, angle, and tactical sector between the player and boss. The judges additionally receive the boss identity, available skill cards, action history, and selected action, but not the generated video. Each skill card specifies its semantic role and effective range, enabling the judges to determine whether the selected action belongs to the boss and is compatible with the current spatial context. For all methods, we evaluate decision quality over the available skill set at each decision point. Boss actions are recovered from generated frames, and the selected skill is assessed against the current gameplay context. Automatic melee actions and continued skill animations are excluded. The prompt is shown in Fig. \ref{fig:decision_quality_prompt}.

 \begin{figure}[t]
      \centering
  \begin{tcolorbox}[
      width=\linewidth,
      colback=white,
      colframe=gray!35,
      colbacktitle=gray!25,
      coltitle=black,
      title=\textbf{Evaluation Prompt (Decision Quality)},
      fonttitle=\normalsize\bfseries,
      boxrule=0.8pt,
      arc=3mm,
      left=1mm,
      right=1mm,
      top=0.8mm,
      bottom=0.8mm
  ]
  You are evaluating the boss decision quality over a 60-second
  gameplay rollout.

  You are given the boss identity, its skill descriptions, and a sequence of
  12 five-second decision steps. Each step contains the predicted combat
  geometry---including distance, orientation, and tactical sector---and the
  boss action recognized in the generated video.

  No video is shown. Judge only from the structured trajectory and skill
  descriptions.

  For every step, evaluate:

  \begin{enumerate}[
      leftmargin=6mm,
      label=\arabic*.,
      itemsep=0pt,
      topsep=0pt,
      parsep=0pt
  ]
      \item \textbf{action\_validity (0 or 1):}\\
      Whether the selected action reasonably matches
      the current distance, orientation, and tactical role.

      \item \textbf{sequence\_fit (0--5):}\\
      Whether the action contributes to a coherent full-rollout strategy,
      including appropriate adaptation, skill variety, and avoidance of
      excessive repetition.
  \end{enumerate}

  \textbf{Return one JSON object only}, containing one entry for each step
  from 0 through 11.

  \end{tcolorbox}
      \caption{Condensed text-only LLM prompt for Decision Quality.}
      \label{fig:decision_quality_prompt}
  \end{figure}
  
\noindent\textbf{Visual Understanding.}
We separately evaluate the perception modules that provide observations to the agentic control layer. AttackNet is evaluated on 5,779 held-out intervals using Macro-F1 and Accuracy for player- and boss-hit detection. Macro-F1 averages the F1 scores of the two hit signals, while Accuracy requires both predictions to be correct. SituationNet is evaluated on 5,576 valid intervals. Relative distance and angle are measured using MAE, where distance is computed in each game's native coordinate system and angular error uses the shortest circular difference. We additionally report classification accuracy for three spatial indicators: \textit{Front-Cone}, which identifies whether the target is ahead with $|\theta|\leq20^\circ$; \textit{Decision-Zone}, which categorizes the target as front, side, or behind, with behind defined as $|\theta|\geq90^\circ$; and \textit{Tactical-Sector}, which further distinguishes the side region into left and right.

\subsection{More Experimental Results}
\label{app:exp_results}
\noindent\textbf{Detailed Decision Quality Analysis.} We further present a comprehensive boss-level analysis to provide a richer evaluation of boss decision quality across different control and state-modeling settings. Action Validity measures whether each selected boss action matches the current spatial context, and is reported as the percentage of valid actions. Sequence Fit scores the temporal coherence of the complete decision sequence from 0 to 5, considering adaptation, diversity, and excessive repetition. Ours Preference compares matched rollouts using an equal-weight combination of normalized Action Validity and Sequence Fit, and reports how often GameDirector obtains the higher score.

As shown in Tab.~\ref{tab:decision_quality_breakdown}, explicit boss control in Settings (2) and (4) generally improves both action validity and sequence
fit over implicit behavior modeling. Predictive states also benefit implicit control, but provide only limited additional gains when explicit commands are
already available. Nevertheless, GameDirector consistently performs best across all three bosses and under both judges. It achieves 88.4\%/93.0\% overall action validity and 4.34/4.66 sequence fit under GPT-5.5/Gemini-3.1-Pro, outperforming the strongest baseline by 39.9\% and 45.8\% in action validity, respectively. The boss-level results show particularly strong performance for Golden Bird, while substantial gains are also maintained for Mutant Soldier and Plague Leader. Moreover, GameDirector is preferred in 89.8\%--95.8\% of matched rollouts against the four baseline settings. These results demonstrate that closed-loop, situation-aware planning produces more tactically appropriate, adaptive, and temporally coherent boss behavior than implicit generation or direct language-based control.

 \begin{table*}[t]
    \centering
    \caption{
    Boss-level breakdown of decision quality across the five rollout settings in Game N.
    Action Validity is reported as a percentage, Sequence Fit is scored from 0 to 5,
    and Ours Pref. denotes the percentage of matched rollouts in which GameDirector
    outperforms the baseline setting. Results are mean $\pm$ standard deviation
    over three evaluations.
    }
    \label{tab:decision_quality_breakdown}

    \setlength{\tabcolsep}{4pt}
    \renewcommand{\arraystretch}{1.08}

    \resizebox{\textwidth}{!}{
    \begin{tabular}{ll|ccc|ccc}
    \toprule
    & & \multicolumn{6}{c}{\textbf{Decision Quality}} \\
    \cmidrule(lr){3-8}
    & & \multicolumn{3}{c|}{\textbf{GPT-5.5}}
    & \multicolumn{3}{c}{\textbf{Gemini-3.1-Pro}} \\
    \cmidrule(lr){3-5}
    \cmidrule(lr){6-8}

    \textbf{Setting}
    & \textbf{Boss}
    & \textbf{Action Valid. (\%) $\uparrow$}
    & \textbf{Seq. Fit $\uparrow$}
    & \textbf{Ours Pref. (\%) $\uparrow$}
    & \textbf{Action Valid. (\%) $\uparrow$}
    & \textbf{Seq. Fit $\uparrow$}
    & \textbf{Ours Pref. (\%) $\uparrow$} \\
    \midrule

    \multirow{4}{*}{\shortstack[l]{(1) Implicit\\No State}}
    & Golden Bird
    & $54.1_{\pm 1.6}$ & $2.51_{\pm 0.01}$ & $100.0_{\pm 0.0}$
    & $49.1_{\pm 1.1}$ & $2.62_{\pm 0.05}$ & $99.0_{\pm 1.7}$ \\
    & Mutant Soldier
    & $14.3_{\pm 0.7}$ & $1.70_{\pm 0.02}$ & $93.9_{\pm 0.0}$
    & $13.5_{\pm 0.3}$ & $1.61_{\pm 0.16}$ & $93.9_{\pm 0.0}$ \\
    & Plague Leader
    & $13.4_{\pm 1.2}$ & $1.68_{\pm 0.04}$ & $91.7_{\pm 1.8}$
    & $14.4_{\pm 0.6}$ & $1.70_{\pm 0.13}$ & $90.6_{\pm 0.0}$ \\
    \cmidrule(lr){2-8}
    & \textbf{Overall}
    & $28.1_{\pm 0.9}$ & $1.98_{\pm 0.02}$ & $\mathbf{95.3_{\pm 0.6}}$
    & $26.4_{\pm 0.2}$ & $2.00_{\pm 0.11}$ & $\mathbf{94.6_{\pm 0.6}}$ \\

    \midrule

    \multirow{4}{*}{\shortstack[l]{(2) Explicit\\No State}}
    & Golden Bird
    & $77.6_{\pm 1.3}$ & $3.13_{\pm 0.01}$ & $97.1_{\pm 0.0}$
    & $71.5_{\pm 0.6}$ & $3.19_{\pm 0.06}$ & $94.6_{\pm 0.8}$ \\
    & Mutant Soldier
    & $32.5_{\pm 2.4}$ & $2.27_{\pm 0.06}$ & $91.2_{\pm 0.0}$
    & $30.3_{\pm 0.8}$ & $2.27_{\pm 0.05}$ & $94.1_{\pm 0.0}$ \\
    & Plague Leader
    & $28.9_{\pm 1.4}$ & $2.27_{\pm 0.04}$ & $86.5_{\pm 1.8}$
    & $33.2_{\pm 0.9}$ & $2.31_{\pm 0.06}$ & $85.9_{\pm 1.6}$ \\
    \cmidrule(lr){2-8}
    & \textbf{Overall}
    & $48.5_{\pm 0.9}$ & $2.60_{\pm 0.02}$ & $\mathbf{91.7_{\pm 0.6}}$
    & $47.2_{\pm 0.3}$ & $2.64_{\pm 0.02}$ & $\mathbf{91.7_{\pm 0.6}}$ \\

    \midrule

    \multirow{4}{*}{\shortstack[l]{(3) Implicit\\Predictive State}}
    & Golden Bird
    & $46.2_{\pm 0.4}$ & $2.41_{\pm 0.03}$ & $100.0_{\pm 0.0}$
    & $42.5_{\pm 0.7}$ & $2.69_{\pm 0.08}$ & $99.5_{\pm 0.8}$ \\
    & Mutant Soldier
    & $15.6_{\pm 1.4}$ & $1.78_{\pm 0.06}$ & $97.5_{\pm 0.8}$
    & $14.8_{\pm 0.0}$ & $1.79_{\pm 0.09}$ & $96.6_{\pm 0.8}$ \\
    & Plague Leader
    & $17.6_{\pm 0.4}$ & $1.87_{\pm 0.01}$ & $89.6_{\pm 1.8}$
    & $20.4_{\pm 0.4}$ & $1.97_{\pm 0.09}$ & $86.5_{\pm 1.8}$ \\
    \cmidrule(lr){2-8}
    & \textbf{Overall}
    & $27.1_{\pm 0.4}$ & $2.04_{\pm 0.03}$ & $\mathbf{95.8_{\pm 0.3}}$
    & $26.6_{\pm 0.3}$ & $2.17_{\pm 0.04}$ & $\mathbf{94.3_{\pm 1.2}}$ \\

    \midrule

    \multirow{4}{*}{\shortstack[l]{(4) Explicit\\Predictive State}}
    & Golden Bird
    & $73.1_{\pm 2.4}$ & $3.12_{\pm 0.02}$ & $98.0_{\pm 3.4}$
    & $66.7_{\pm 0.8}$ & $3.33_{\pm 0.07}$ & $86.8_{\pm 3.9}$ \\
    & Mutant Soldier
    & $28.4_{\pm 0.5}$ & $2.18_{\pm 0.02}$ & $90.2_{\pm 1.7}$
    & $28.1_{\pm 0.5}$ & $2.17_{\pm 0.09}$ & $93.6_{\pm 2.2}$ \\
    & Plague Leader
    & $31.6_{\pm 2.3}$ & $2.41_{\pm 0.01}$ & $88.5_{\pm 3.6}$
    & $38.0_{\pm 0.3}$ & $2.42_{\pm 0.06}$ & $89.1_{\pm 3.1}$ \\
    \cmidrule(lr){2-8}
    & \textbf{Overall}
    & $44.3_{\pm 1.0}$ & $2.58_{\pm 0.01}$ & $\mathbf{92.3_{\pm 0.6}}$
    & $44.6_{\pm 0.1}$ & $2.65_{\pm 0.03}$ & $\mathbf{89.8_{\pm 1.5}}$ \\

    \midrule

    \multirow{4}{*}{\shortstack[l]{\textbf{GameDirector}\\\textbf{(Ours)}}}
    & Golden Bird
    & $\mathbf{100.0_{\pm 0.0}}$ & $\mathbf{4.52_{\pm 0.02}}$ & --
    & $\mathbf{99.5_{\pm 0.9}}$ & $\mathbf{4.72_{\pm 0.01}}$ & -- \\
    & Mutant Soldier
    & $\mathbf{80.4_{\pm 0.0}}$ & $\mathbf{4.20_{\pm 0.02}}$ & --
    & $\mathbf{92.9_{\pm 0.0}}$ & $\mathbf{4.66_{\pm 0.06}}$ & -- \\
    & Plague Leader
    & $\mathbf{80.2_{\pm 3.1}}$ & $\mathbf{4.23_{\pm 0.06}}$ & --
    & $\mathbf{82.0_{\pm 1.6}}$ & $\mathbf{4.54_{\pm 0.10}}$ & -- \\
    \cmidrule(lr){2-8}
    & \textbf{Overall}
    & $\mathbf{88.4_{\pm 0.7}}$ & $\mathbf{4.34_{\pm 0.01}}$ & --
    & $\mathbf{93.0_{\pm 0.6}}$ & $\mathbf{4.66_{\pm 0.04}}$ & -- \\

    \bottomrule
    \end{tabular}
    }
    \end{table*}

 \begin{table}[t]
    \centering
    \small
    \caption{Deployment configuration and steady-state inference speed. Component
    latencies are reported for the four-GPU deployment. Throughput is averaged
    over ten measured rollouts after one excluded warm-up rollout.}
    \label{tab:deployment_inference_speed}
    \begin{tabular}{llrr}
    \toprule
    Component & Device or Deployment & Mean Latency / Speed & Calls \\
    \midrule
    DiT world model & GPU 0 & 225.060 ms/cell & 220 \\
    AttackNet & GPU 1 & 15.108 ms/call & 221 \\
    SituationNet & GPU 2 & 17.749 ms/call & 22 \\
    Gemma-4-E2B-it & GPU 3 & 2230.264 ms/request & 11 \\
    \midrule
    Agentic rollout & Four GPUs & 17.449 FPS & 10 \\
    Agentic rollout & Two GPUs & 17.277 FPS & 10 \\
    Agentic rollout & One GPU & 16.729 FPS & 10 \\
    \midrule
    DiT-cell ceiling & Four GPUs & 17.773 FPS & -- \\
    DiT-cell ceiling & Two GPUs & 17.524 FPS & -- \\
    DiT-cell ceiling & One GPU & 17.026 FPS & -- \\
    \midrule
    Ceiling utilization & Four GPUs & 98.18\% & -- \\
    Ceiling utilization & Two GPUs & 98.59\% & -- \\
    Ceiling utilization & One GPU & 98.25\% & -- \\
    \bottomrule
    \end{tabular}
  \end{table}
  
 \noindent\textbf{Deployment Setting and Inference Speed.} As shown in Tab. \ref{tab:deployment_inference_speed}, we evaluate the agentic rollout pipeline under four-, two-, and one-GPU deployments. In the four-GPU setting, the distilled Stage-3 world model runs on GPU~0 with three denoising steps at a resolution of $480\times832$, while AttackNet, SituationNet, and Gemma-4-E2B-it are placed on GPUs~1, 2, and 3, respectively. In the two-GPU setting, the world model exclusively occupies one GPU, while the other three models share the second GPU. In the one-GPU setting, all four models share a single GPU. The LLM starts planning at the second simulated second of each five-second tick, providing a three-second lookahead window. This completely hides its approximately 2.2-second inference latency in all three settings. For efficiency, AttackNet inference overlaps with generation of the next DiT cell. When an attack reduces either character's HP to zero, generation directly switches to the corresponding death prompt. We measure ten consecutive steady-state rollouts after excluding one warm-up rollout, generating 800 frames in total. The measurement includes planning, perception, combat logic, world-model generation, and HUD rendering. The four-, two-, and one-GPU deployments achieve 17.449, 17.277, and 16.729 FPS, respectively, corresponding to 98.18\%, 98.59\%, and 98.25\% of their measured DiT-cell throughput ceilings. Thus, even when all components share one GPU, the agentic framework retains more than 98\% of the available generation throughput, demonstrating that its planning, perception, and control pipeline introduces minimal runtime overhead.

\subsection{Baseline Details}
\label{app:baseline_details}

The cited methods in Tab. \ref{tab:main_results} are used only as representative references for their state-modeling and boss-control paradigms, rather than being reproduced exactly. To balance architectural fidelity and fair comparison, all settings retain the same Wan2.2-TI2V-5B backbone and use identical training data, player controls, evaluation samples, and generation configurations. We introduce only the minimal modifications required to instantiate each combination of state modeling and boss control.

\noindent\textbf{Setting (1): No State Modeling with Implicit Boss Control.}
Following the representative design of Matrix-Game~\citep{he2025matrix}, this setting treats gameplay generation as visual prediction without explicitly representing numerical game states. The model receives the initial frame and player actions, while boss behavior is learned implicitly from the training videos. During a one-minute rollout, only the final generated frame is propagated between consecutive five-second segments; no state trajectory or boss-action command is provided.

\noindent\textbf{Setting (2): No State Modeling with Explicit Boss Control.} Inspired by the language-conditioned control used in ReactiveGWM~\citep{wang2026reactivegwm}, this setting retains the same state-free visual architecture as Setting (1), but additionally provides a text condition for each segment. The text specifies the boss identity, intended actions, and high-level strategy, sampled from training fights involving the same boss and scene. This enables direct boss control without introducing numerical state prediction. However, the commands are predetermined and do not adapt to the generated gameplay.

\noindent\textbf{Setting (3): Predictive State Modeling with Implicit Boss Control.} Following the predictive-state paradigm represented by WildWorld~\citep{li2026wildworld}, this setting augments Setting (1) with a state branch that jointly predicts player HP, boss HP, and the boss skill meter alongside the video. The final predicted frame and state are both propagated to the next segment, forming an autoregressive trajectory over the complete rollout. Boss behavior remains implicit because the model receives no boss-action or strategy description.

\noindent\textbf{Setting (4): Predictive State Modeling with Explicit Boss Control.} Following the state-aware generation design represented by StatePlay~\citep{lin2026stateplay}, this setting combines predictive state modeling with explicit text-based boss control. It jointly generates video and combat states as in Setting (3), while also receiving the boss-action and strategy descriptions used in Setting (2). Although this configuration provides both state information and direct language guidance, its boss commands are fixed before generation rather than selected online according to the evolving gameplay situation.

\subsection{More Visualization Results}
\label{app:subsec_more_visual}
\begin{figure*}[t]
\centering
    \includegraphics[width=\textwidth]{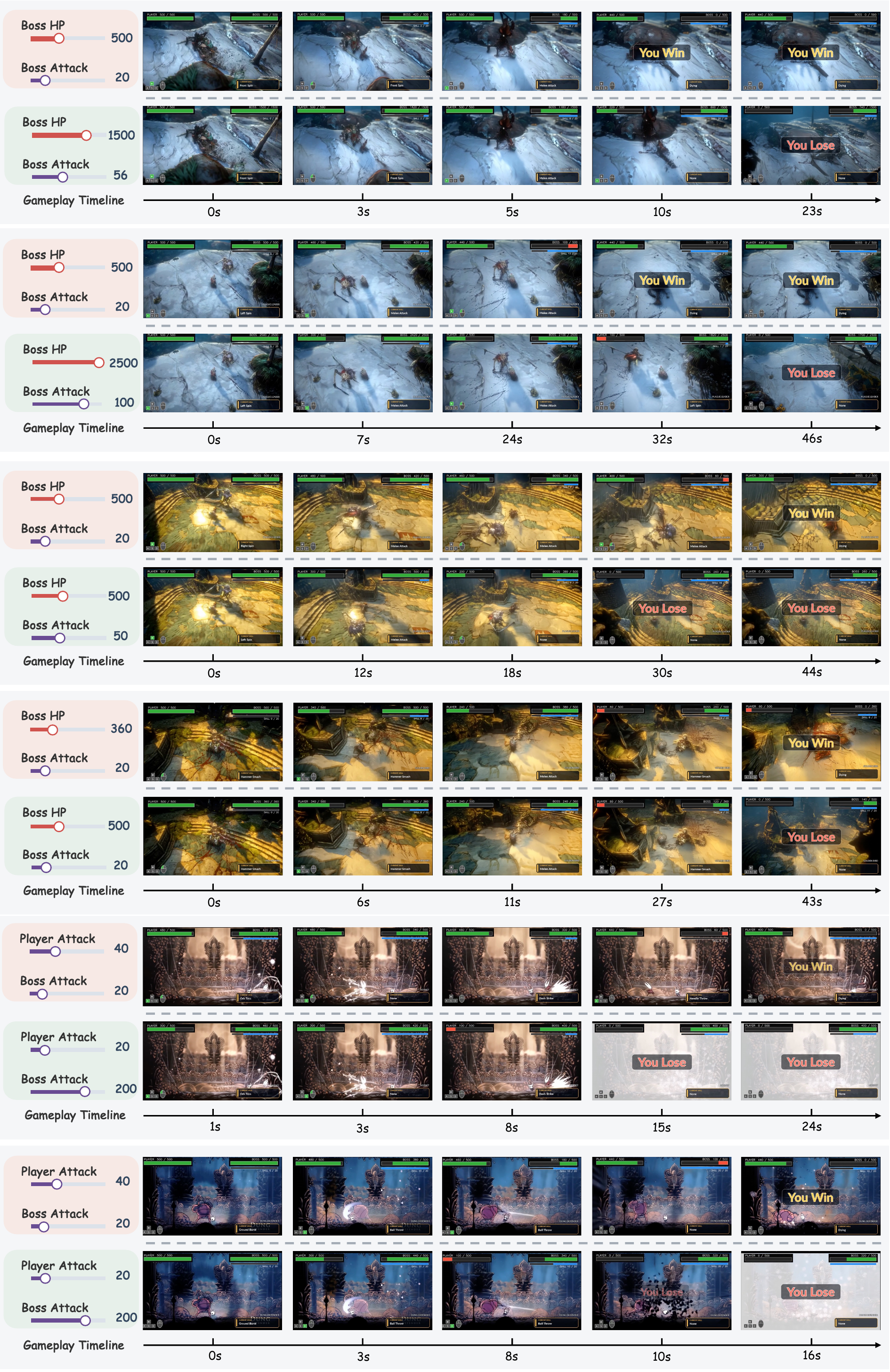}
    \caption{
   Additional visualization of player-configurable initialization. Varying boss HP, boss attack strength, and player attack strength results in distinct gameplay trajectories and outcomes, consistently reflecting the configured game states over long-horizon rollouts.
    }
    \label{fig:app_vis}
\end{figure*}

We provide additional visualization results of player-configurable initialization in Fig.~\ref{fig:app_vis}. By varying the initial boss HP, boss attack strength, and player attack strength, GameDirector produces distinct gameplay trajectories that consistently reflect the configured states. Lower boss HP or attack strength makes the encounter easier for the player, while increasing boss HP or attack strength leads to longer and more challenging combat. Similarly, modifying the player's attack strength directly affects combat progression and final outcomes. These results further demonstrate that player-defined configurations are reliably reflected throughout long-horizon gameplay, allowing players to customize the game experience according to their preferences.

\label{app:baseline_details}

\end{document}